\documentclass[10pt,twocolumn,letterpaper]{article}

\usepackage[pagenumbers]{cvpr} 
\usepackage{booktabs}
\usepackage{url}
\usepackage{amsmath,amssymb,graphicx}
\usepackage[pagebackref,breaklinks,colorlinks]{hyperref}

\usepackage{lipsum} 

\usepackage{enumitem}
\usepackage{booktabs} 
\usepackage{amsmath}
\usepackage{cite}
\usepackage{cleveref}
\usepackage{subcaption}
\usepackage{xcolor}
\usepackage{soul} 

\usepackage[switch]{lineno}

\author{\IEEEauthorblockN{Erez Yosef, Raja Giryes}\\
    \IEEEauthorblockA{Tel-Aviv University, Israel}
    }

\begin{document}

\title{DifuzCam: Replacing Camera Lens with\\ a Mask and a Diffusion Model}

\author{Erez Yosef\\
Tel Aviv University, Israel\\
{\tt\small erez.yo@gmail.com}
\and
Raja Giryes\\
Tel Aviv University, Israel\\
{\tt\small raja@tauex.tau.ac.il}
}

\maketitle
\begin{abstract}
The flat lensless camera design reduces the camera size and weight significantly. In this design, the camera lens is replaced by another optical element that interferes with the incoming light. The image is recovered from the raw sensor measurements using a reconstruction algorithm. Yet, the quality of the reconstructed images is not satisfactory.
To mitigate this, we propose utilizing a pre-trained diffusion model with a control network and a learned separable transformation for reconstruction. This allows us to build a prototype flat camera with high-quality imaging, presenting state-of-the-art results in both terms of quality and perceptuality. We demonstrate its ability to leverage also textual descriptions of the captured scene to further enhance reconstruction.
Our reconstruction method which leverages the strong capabilities of a pre-trained diffusion model can be used in other imaging systems for improved reconstruction results.
\end{abstract}

\section{Introduction}\label{sec:introduction}

Cameras have become very popular and common in recent years, especially in small handheld devices. Despite that, reducing the size of the camera remains a difficult problem since a camera requires lenses and optical elements to get a high-quality image. Flat cameras \cite{salman2015flatcam} is a computational photography method to reduce camera size by replacing the camera lens with a diffuser, namely, an amplitude mask placed very close to the sensor. Thus, the image on the sensor consists of multiplexed projections of the scene reflections on all the sensors area such that the captured image is not visually understandable. Using a computational algorithm, the scene image can be retrieved. Yet, while achieving massive camera size reduction, reconstructing high-quality images from flat camera measurements is an ill-posed task and hard to achieve. 

Previous approaches tried to reconstruct the scene image using different techniques including direct optimization \cite{salman2015flatcam} and deep learning \cite{khan2019towardsflatnet}.
Despite these attempts, the resulting images are not of sufficient quality. High-quality image reconstruction from flat camera measurements is still not solved and better algorithms are required to reproduce better images.

\begin{figure}
    \setlength\tabcolsep{2pt} 
    \centering
    \begin{subfigure}[b]{0.37\linewidth}
    \includegraphics[width=\linewidth]{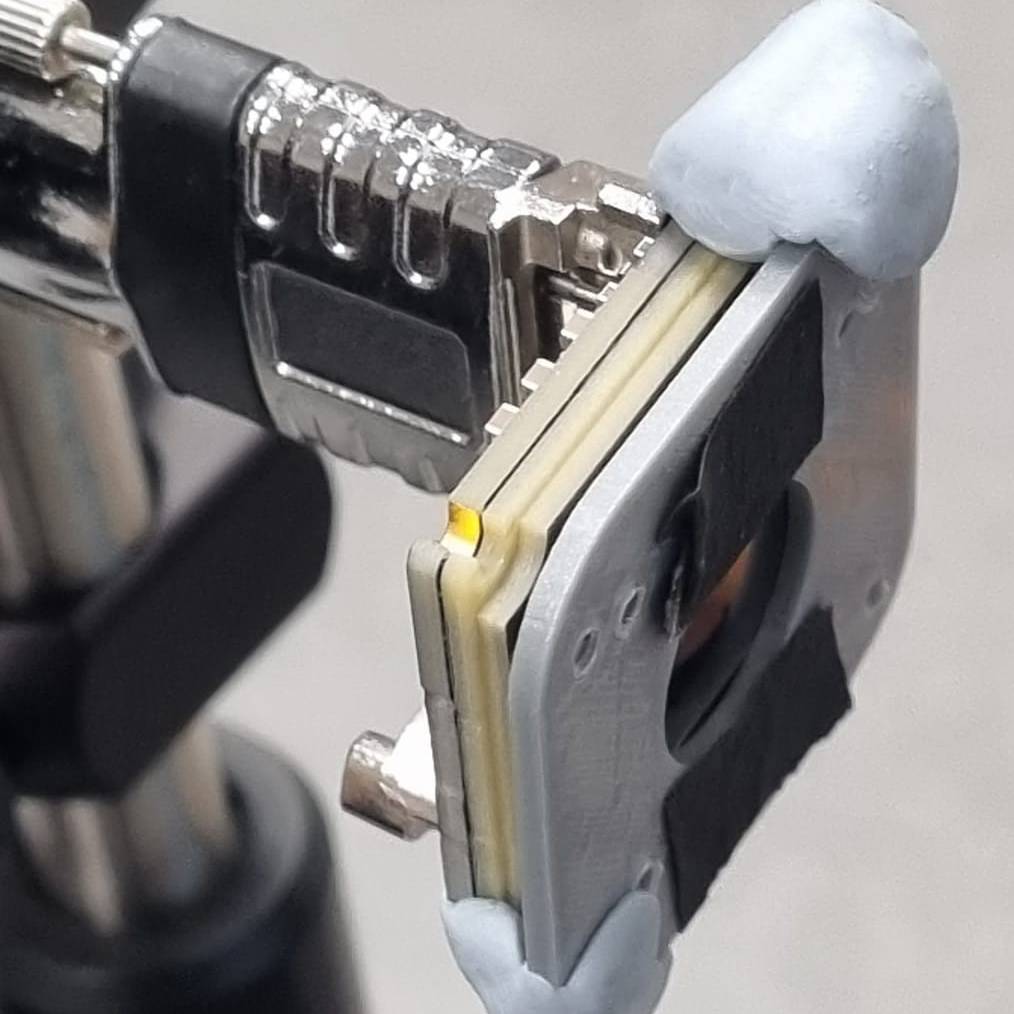}
         \caption{\footnotesize Prototype flat camera}
         \label{fig:teaser_a}
    \end{subfigure}
    \hspace{0.1cm}
        \begin{subfigure}[b]{0.45\linewidth}
    \includegraphics[width=\linewidth]{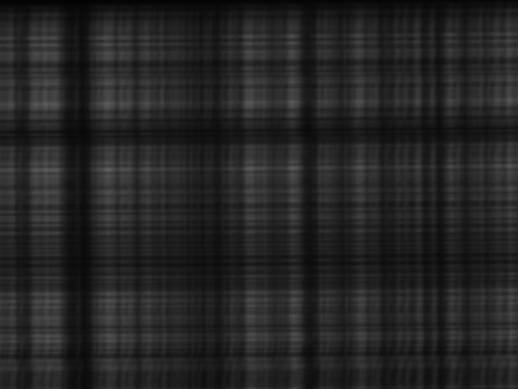}
         \caption{\footnotesize Captured image}
         \label{fig:teaser_b}
    \end{subfigure}
    \begin{subfigure}[b]{0.37\linewidth}
    \includegraphics[width=\linewidth]{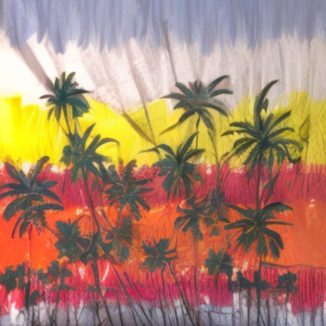} 
         \caption{\footnotesize Reconstruction}
         \label{fig:teaser_c}
         \end{subfigure}
         \hspace{0.2cm}
    \begin{subfigure}[b]{0.37\linewidth}
    \includegraphics[width=\linewidth]{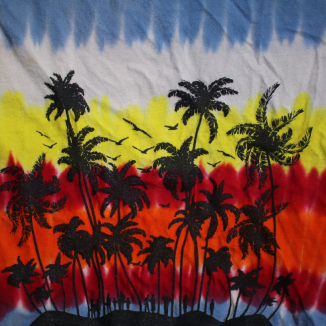}
         \caption{\footnotesize Reference}
         \label{fig:teaser_d}
    \end{subfigure}
    \caption{Using (a) our prototype flat camera, (b) a measurement image is captured that is not visually understandable. (c) An image is reconstructed from the measurements using our text-guided approach. Compare to (d) the reference image captured with a regular camera. (see \Cref{fig:prototype_result_real} for details).}
    \vspace{-0.15in}
    \label{fig:teaser}
\end{figure}

\begin{figure*}
    \setlength\tabcolsep{4pt} 
    \centering
    \includegraphics[width=0.8\textwidth]{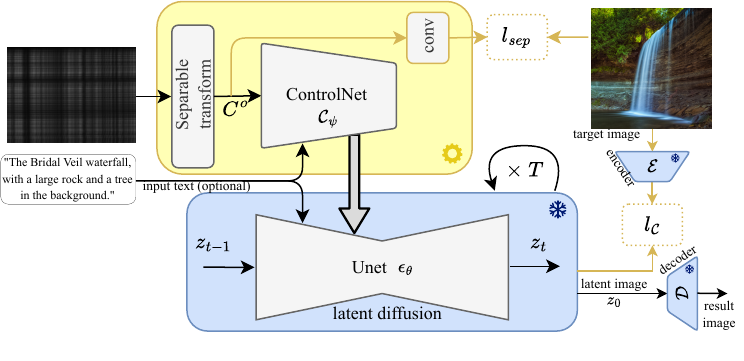}
    \vspace{-0.1in}
    \caption{\textbf{DifuzCam Proposed Method.} A flat lensless camera measurements are inputted to a separable linear transformation followed by a ControlNet adapter to control the image generation process by a pre-trained latent diffusion model (LDM). Using text guidance for the reconstruction process is optional in our method. On training, the {\textbf{yellow}} blocks weights are optimized, and {\textbf{orange}} paths are the training losses, while the {\textbf{blue}} blocks weights are pre-trained and fixed. The dataset for training and tests was captured using our prototype flat camera by projecting images onto a screen. We present an additional loss $l_{sep}$ in addition to the diffusion training loss ($l_{\mathcal{C}}$) for better convergence. Reconstructed image is achieved by $T$ iterative diffusion steps and decoding from latent space to pixel space.}
    \label{fig:overview}
    \vspace{-0.1in}
\end{figure*}

We propose \textit{DifuzCam}, a novel strategy for flat camera image reconstruction using a strong image prior that relies on a pre-trained diffusion model \cite{rombach2022high, croitoru2023diffusion}. An overview diagram of DifuzCam is presented in \Cref{fig:overview}. Using a pre-trained image generation model, which is trained on a huge amount of images, we utilize a strong prior for natural images perceptually. This facilitates reconstructing state-of-the-art quality images from flat camera measurements (\Cref{fig:teaser_b}). Moreover, we leveraged the diffusion model text-guidance property for generation to improve even further image reconstruction through a description of the captured scene from the photographer. 

\definecolor{cy}{rgb}{1, 1, 0.71}
\definecolor{cb}{rgb}{0.86, 0.93, 1} 
\definecolor{co}{rgb}{0.97, 0.835, 0.435} 
\newcommand{\hlc}[2][yellow]{{%
    \colorlet{foo}{#1}%
    \sethlcolor{foo}\hl{#2}}%
}

We used stable diffusion \cite{rombach2022high} and trained a ControlNet \cite{zhang2023adding} for our flat camera images captured by our flat camera prototype (\Cref{fig:teaser_a}). We present qualitative and quantitative evaluations of our proposed method using both our camera prototype and existing datasets from prior work to perform a comparison.

The method we present improves the reconstruction from flat camera measurements and thus, makes these cameras more practical. Additionally, our proposed approach can be modified for other imaging systems and setups. It leverages the diffusion model's strong prior and text guidance capability to improve the reconstructed image quality.

To conclude, our main contributions are: (i) A novel computational photography method based on a lensless flat camera and a reconstruction algorithm based on diffusion model image prior; (ii) State-of-the-art results for flat camera reconstruction quality in all evaluated metrics; (iii) A novel approach to using text guidance to improve imaging results of an optical system; and (iv) We present a deep control network with intermediate separable loss for improving convergence and results.

\section{Related Work}
\label{sec:related_work}
Lensless imaging has garnered significant attention in recent years since it offers a more compact, light, and cost-effective imaging system. 
Different approaches to lensless camera design were proposed, such as static amplitude mask \cite{salman2015flatcam}, modulation with LCD \cite{huang2013lensless}, Fresnel zone aperture (FZA) with SLM \cite{deweert2014lensless,shimano2018lensless, nakamura2016lensless}, phase mask \cite{boominathan2020phlatcam, antipa2018diffusercam}, programmable devices \cite{zomet2006lensless, miller2020particle, zheng2021simple, hua2020sweepcam}, and more as described in the following review \cite{boominathan2022recent}. 

In this paper, we rely on the FlatCam design \cite{salman2015flatcam}. It employs a static amplitude mask placed near the sensor. The mask pattern is designed in a separable manner such that the imaging linear model can be simplified to a separable operation:
\begin{equation}\label{eq:imaging_model}
    Y = \Phi_lX\Phi_r,
\end{equation}
where $\Phi_L$ and $\Phi_R$ are the separable operations on the image’s rows and columns, respectively. $X$ is the scene intensity and $Y$ is the sensor measurements.

In this case, the measurements of the camera consist of multiplexed projections of the scene reflectance such that reconstruction becomes an ill-posed problem. Prior methods for image reconstruction from flat camera measurements were designed as an optimization problem \cite{salman2015flatcam, hua2020sweepcam, zheng2021simple}. Such model-based methods are heavily dependent on the imaging model and rely on accurate calibration. Thus, they were very limited and the reconstruction quality was low.

To overcome some of these limitations, data-driven algorithms were proposed using deep learning to reconstruct high-quality images from multiplexed measurements. For example, deep learning was used to optimize the alternating direction method of multipliers (ADMM) parameters and reconstruct the image \cite{radford2021learning}. FlatNet \cite{khan2020flatnet, khan2019towardsflatnet} used learned separable transform followed by a Unet \cite{ronneberger2015u} architecture which perceptually enhances the resulting image in a generative adversarial network (GAN) approach. 
 GAN is also presented for lensless point spread function (PSF) estimation and robust reconstruction \cite{rego2021robust}.
 In \cite{wu2021dnn}, a Unet architecture followed by a deep back-projection network (DBPN) \cite{haris2018deep} was proposed for FZA lensless camera image reconstruction. 
 To address the mismatch between the ideal imaging model and the real-world model, the work in \cite{li2023mwdns} presented multi-Wiener deconvolution networks in multi-scale feature spaces.
The research in \cite{liu2023lensless} presented a multi-level image restoration with the pix2pix generative adversarial network.
A recent work claimed improved results by incorporating the transformer architecture into the reconstruction model \cite{pan2022image}.

\noindent \textbf{Diffusion models}  \cite{ho2020denoising,nichol2021improved,croitoru2023diffusion} are generative models that have gained a lot of popularity in recent years due to their impressive ability to learn the natural image distribution which are used for image generation \cite{rombach2022high, nichol2021glide, dhariwal2021diffusion}, segmentation \cite{amit2021segdiff, baranchuk2021label}, inpainting \cite{lugmayr2022repaint}, image super-resolution \cite{saharia2022image}, and general image reconstruction \cite{kawar2022denoising, saharia2022palette, batzolis2021conditional, abu2022adir, zhu2023denoising, fei2023generative,croitoru2023diffusion}. 
The diffusion model is a parameterized Markov chain that produces samples of a certain data distribution after a predefined $T$ number of steps. In each step of the forward diffusion process, a Gaussian noise is added to the data, such that after a large number of steps, the sample is mapped to an isotropic Gaussian. In the reverse step, a deep neural network is trained to denoise the clean image from the noisy sample. To sample an image, we apply $T$ steps in the reverse direction starting with pure Gaussian noise and denoising it until getting a clean natural image from the learned distribution of the data.

In the context of low-level image restoration, diffusion models were recently used for image restoration of linear inverse problems \cite{kawar2022denoising, chung2023prompt, delbracio2023inversion}, spatially-variant noise removal \cite{pearl2023svnr} and low-light image enhancement and denoising \cite{yi2023diff, torem2023complex}.
The diffusion method was also presented for a low-light text recognition task \cite{nguyen2023diffusion}. The use of text guidance for conventional imaging reconstruction and enhancement was recently proposed \cite{kim2023regularization,qi2023tip, yosef2023tell}. For non-conventional imaging, such as a flat camera, adopting diffusion models for the recovery process is not trivial and additional adjustments and considerations are required. This work bridges this gap and presents a novel approach to integrating diffusion models for this imaging task.

Latent diffusion model (LDM) \cite{rombach2022high} is a method for applying the diffusion process in a latent space instead of the pixel space to get a more computationally efficient model for high-resolution images. In this approach, a pre-trained autoencoder consisting of encoder $\mathcal{E}$ and decoder $\mathcal{D}$ is used to train a diffusion model in a low-dimension (latent) space.

The data samples are denoted as $x \sim q(x)$, where $q$ is the data distribution we learn, and the latent sample is obtained by the encoder as $z = \mathcal{E}(x)$. A denoiser network $\epsilon_\theta$ is trained to predict the added noise (at the backward diffusion step) from the noisy sample $z_t$ at the timestamp $t \in [0,\ldots,T]$.
With an additional text input $y$ for text guidance in image generation, the training loss for the LDM is 
\begin{equation}\label{eq:lldm}
l_{ldm}=\mathbb{E}_{\mathcal{E}(x), \epsilon \sim N(0,\mathbf{I}), t \sim U(0,T)}\Bigl[ ||\epsilon - \epsilon_\theta(z_t, t,y) || \Bigr].
\end{equation}

ControlNet \cite{zhang2023adding} adds an image encoder to control the diffusion process. This control network is initiated as the pre-trained encoder of the diffusion model UNet with additional zero convolutions. The output features of the control network are added to the pre-trained diffusion model features, and due to the zero convolutions, the initial performance of the diffusion model is not degraded. After training, ControlNet weights are optimized such that the added control network features manipulate the diffusion process for the target task. Other methods for controlling the generation process of diffusion models for a specific task were also suggested \cite{mou2024t2i, duan2024tuning}.

\begin{figure}[t]
    \setlength\tabcolsep{2pt} 
    \centering
    \begin{subfigure}[b]{0.375\linewidth}
    \includegraphics[width=\linewidth]{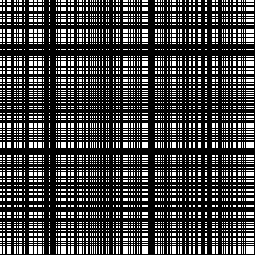}
         \caption{Mask pattern}
    \end{subfigure}
    \hspace{0.3cm}
        \begin{subfigure}[b]{0.5\linewidth}
    \includegraphics[width=\linewidth]{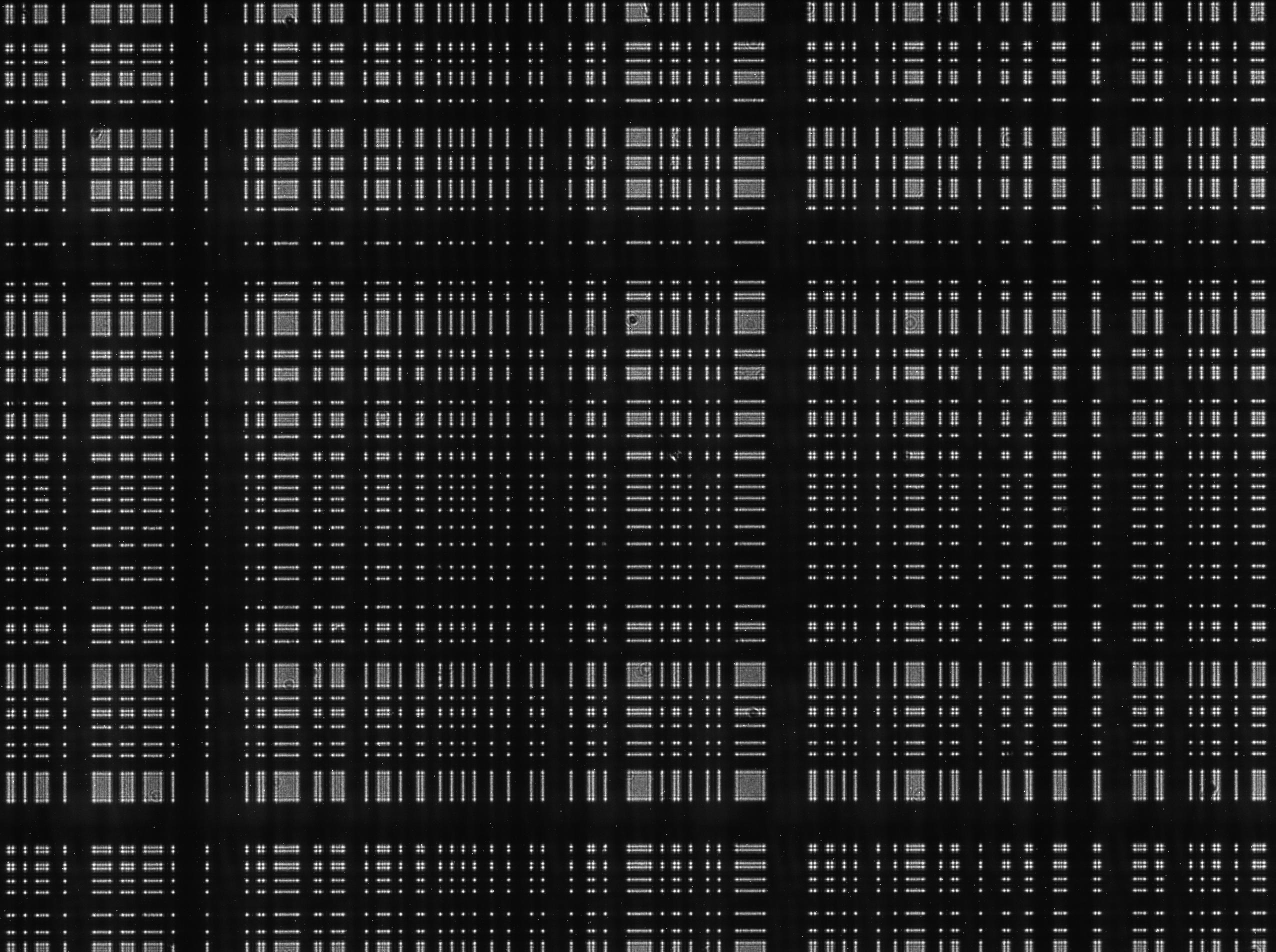}
         \caption{Measured PSF}
    \end{subfigure}

    \caption{\textbf{Optical design.} (a) The designed separable mask which was printed on a chrome plate using lithography. (b) The measured PSF by the prototype camera with the coded mask}
    \vspace{-0.15in}
    \label{fig:psf}
\end{figure}

\section{Proposed Method}
\label{sec:method}
\begin{table}[t]
\renewcommand{\arraystretch}{1.3}
\caption{Evaluation and comparison to prior work. Our method outperforms the compared method FlatNet in all metrics. Employing text guidance improves further more the reconstruction results.}
\centering
\begin{tabular}{lcccc}
\hline
method & PSNR$\uparrow$ & SSIM$\uparrow$ & LPIPS$\downarrow$ & CLIP$\uparrow$\\
\hline\hline
Tikhonov \cite{salman2015flatcam} & 10.98&	0.318&	0.736 & 16.76\\
flatnet-R \cite{khan2019towardsflatnet} & 18.58&	0.487&	0.332 & 21.53\\

flatnet-T \cite{khan2019towardsflatnet} & 18.64	&0.518	&0.322 & 21.68\\
\hline
\multicolumn{3}{@{}l}{\textbf{DifuzCam (our)}}\\
 \footnotesize without text        & \underline{20.25} & 0.601 & 0.242 & 22.81\\
 \footnotesize sampled with text  & 20.20 & \underline{0.604}       & \underline{0.238} & \underline{23.50}\\
 \footnotesize fine-tuned with text      & \textbf{20.43} & \textbf{0.612} & \textbf{0.237} & \textbf{23.53}\\
\hline
\label{tab:flatnet_results}
\end{tabular}
\end{table}

\newcommand{\pfig}[1]{\includegraphics[width=0.18\linewidth]{#1}}
\newcommand{\Eraw}{figs/laion/raw/}
\newcommand{\Eour}{figs/laion/w_caps/} %
\newcommand{\Eournc}{figs/laion/wo_caps/} 

\newcommand{\Egt}{figs/laion/gt/}
\newcommand{\E}{example-image-a}
\begin{figure*}
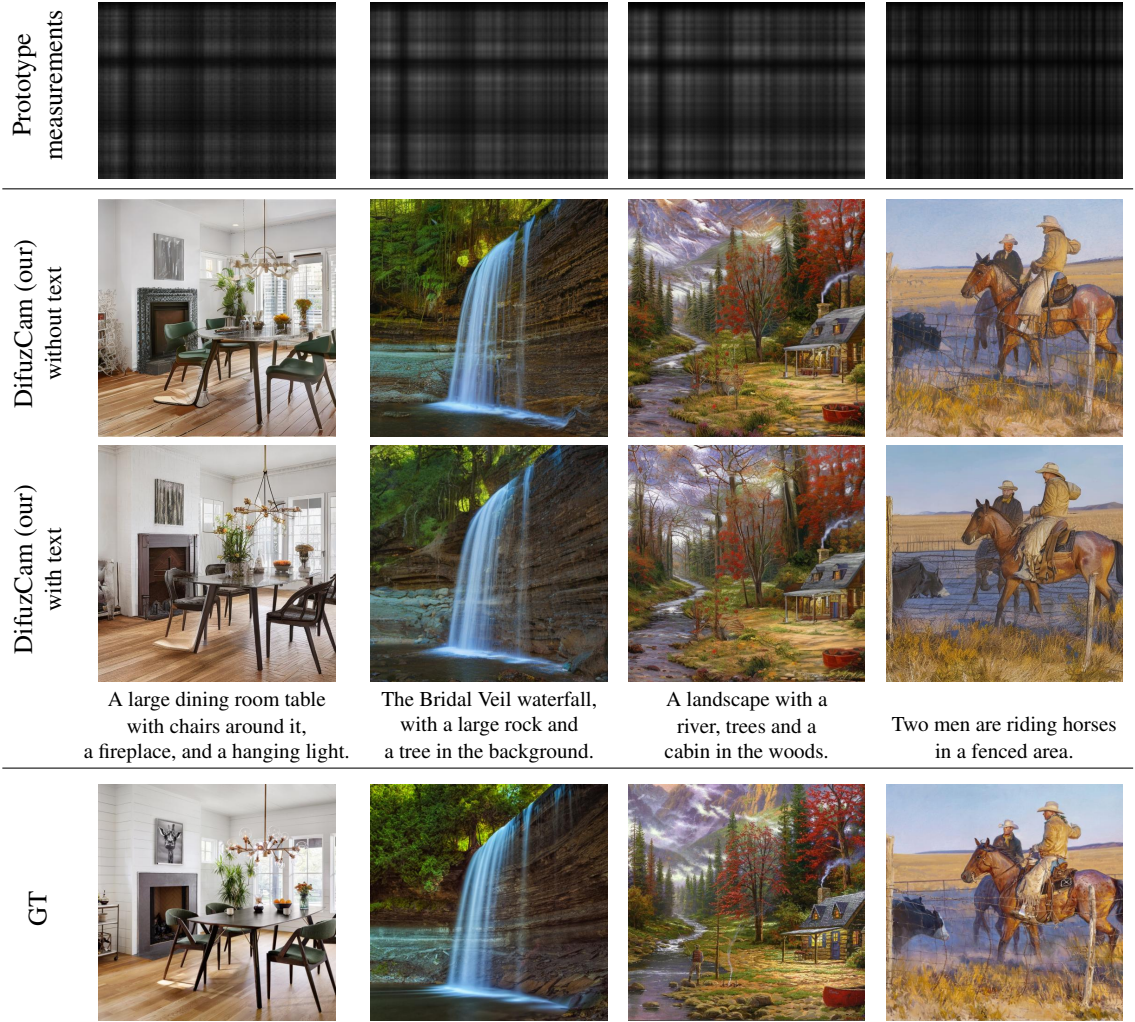

    \setlength\tabcolsep{4pt} 
    \centering
    \begin{tabular}{ccccc}
        \rotatebox{90}{\parbox[c]{2.5cm}{\centering Prototype \\measurements}} & \pfig{\Eraw 9.jpg} & \pfig{\Eraw 10.jpg}& \pfig{\Eraw 18.jpg}& \pfig{\Eraw 32}\\
        \hline \vspace{-8pt} \\
        \rotatebox{90}{\parbox[c]{3cm}{\centering DifuzCam (our) \\ \small without text}} & \pfig{\Eournc validation_8} & \pfig{\Eournc validation_9}& \pfig{\Eournc validation_17}& \pfig{\Eournc validation_31}\\

        \rotatebox{90}{\parbox[c]{3cm}{\centering DifuzCam (our) \\ \small with text}} & \pfig{\Eour validation_8 } & \pfig{\Eour validation_9}& \pfig{\Eour validation_17}& \pfig{\Eour validation_31}\\
          & \shortstack{\footnotesize A large dining room table\\ \footnotesize with chairs around it, \\ \footnotesize a fireplace, and a hanging light.}
          & \shortstack{\footnotesize The Bridal Veil waterfall,\\ \footnotesize with a large rock and \\ \footnotesize a tree in the background.
}
          & \shortstack{\footnotesize A landscape with a \\ \footnotesize river, trees and a \\ \footnotesize  cabin in the woods.}
          & \shortstack{\footnotesize Two men are riding horses \\ \footnotesize in a fenced area.
}\\
          \hline \vspace{-6pt} \\ 
        \rotatebox{90}{\parbox[c]{3cm}{\centering GT}} & \pfig{\Egt targets_8} & \pfig{\Egt targets_9} & \pfig{\Egt targets_17} & \pfig{\Egt targets_31}\\

    \end{tabular}
    \caption{\textbf{Qualitative Results.} Examples of reconstruction results for our captured dataset with the prototype camera we designed.}
    \vspace{-0.1in}
    \label{fig:laion_result}
\end{figure*}

\renewcommand{\pfig}[1]{\includegraphics[width=0.16\linewidth]{#1}}
\newcommand{\Cflatnet}{figs/compare/flatnet/}
\newcommand{\Cour}{figs/compare/our/}
\newcommand{\Cournc}{figs/compare/ournotxt/}

\newcommand{\Cgt}{figs/compare/gt/}
\begin{figure*}[t]
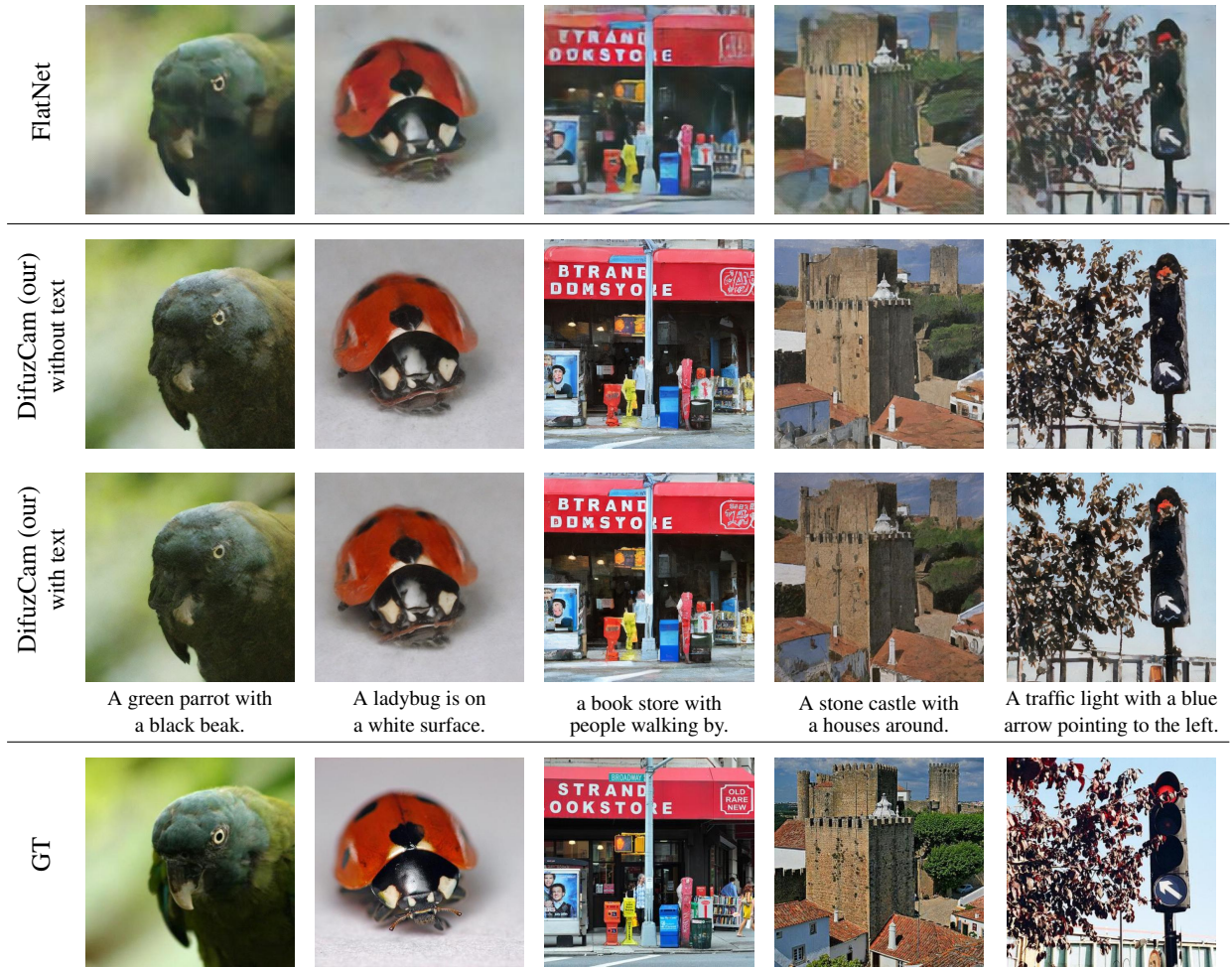

    \setlength\tabcolsep{4pt} 
    \centering
    \begin{tabular}{cccccc}
        \rotatebox{90}{\parbox[c]{3cm}{\centering FlatNet}} & \pfig{\Cflatnet img003} & \pfig{\Cflatnet img025}& \pfig{\Cflatnet img051}& \pfig{\Cflatnet img055}& \pfig{\Cflatnet img076}\\
        \hline
        \rotatebox{90}{\parbox[c]{3cm}{\centering DifuzCam (our) \\ \small without text}} & \pfig{\Cournc validation_2_nc} & \pfig{\Cournc validation_24_nc}& \pfig{\Cournc validation_50_nc}& \pfig{\Cournc validation_54_nc}& \pfig{\Cournc validation_75_nc}\\

        \rotatebox{90}{\parbox[c]{3cm}{\centering DifuzCam (our) \\ \small with text}} & \pfig{\Cour validation_2} & \pfig{\Cour validation_24}& \pfig{\Cour validation_50}& \pfig{\Cour validation_54}& \pfig{\Cour validation_75}\\
          & \shortstack{\footnotesize A green parrot with\\   \footnotesize a black beak.}
          & \shortstack{\footnotesize A ladybug is on \\ \footnotesize a white surface.}
          & \shortstack{ \footnotesize a book store with\\ \footnotesize people walking by.}
          & \shortstack{\footnotesize A stone castle with\\ \footnotesize a houses around.}
          & \shortstack{\footnotesize A traffic light with a blue\\ \footnotesize arrow pointing to the left.}\\
          \hline
        \rotatebox{90}{\parbox[c]{3cm}{\centering GT}} & \pfig{\Cgt targets_2} & \pfig{\Cgt targets_24} & \pfig{\Cgt targets_50} & \pfig{\Cgt targets_54} & \pfig{\Cgt targets_75}\\

    \end{tabular}
    \caption{\textbf{Results Comparison.} We compare the results of our proposed method to the previous method FlatNet-T \cite{khan2020flatnet} on their published data set with their network weights.}
    \vspace{-0.15in}
    \label{fig:flatnet_result}
\end{figure*}

We turn to describe our DifuzCam strategy. The flat camera we use employs a similar implementation to previous works \cite{salman2015flatcam, khan2020flatnet}. A separable pattern obtained from an outer product of M-sequence binary signals in the length of $255$ was used for the amplitude mask. We printed it by lithography on a chrome plate on glass with a thickness of 0.2mm. Each feature in the pattern (i.e. a single pinhole) is of size 25 um. \Cref{fig:psf} shows the mask pattern and the measured camera PSF achieved while the mask is attached to the sensor hot mirror.

The acquired image by a flat camera consists of multiplexed measurements of the light reflected from the scene across the sensor area. The light projections on the sensor create long-range correspondences in the captured images. 
To convert the image from the projections (i.e. "projection space") to the pixel space of the target image we apply a learned separable linear transformation to the measurements.
This transformation is crucial since the input image serves as guidance for the diffusion model process, which was trained and worked in the domain of natural images. The diffusion model is ignorant about the flat camera mask projections on the image, and the control network we use to guide the reconstruction process exhibits better results when operating in the image pixel domain rather than the long-range projections. 

The RAW image captured by the camera (in RGGB Bayer pattern) is split into four color channels: $R$, $G_r$, $G_b$ and $B$. Each channel $C_k$ is linearly transformed as
\begin{equation}\label{eq:separable_transform}
    C^o_k = \phi_l^kC_k\phi_r^k,
\end{equation}
where $k \in [R, G_r, G_b, B]$, each color channel $C_k$ of size $h_i\times w_i$ and $\phi_l \in \mathbb{R}^{h_o\times h_i}$ and $\phi_r \in \mathbb{R}^{w_i\times w_o}$ are two learnable matrices. The output features were stacked onto a single 4-channel tensor $C^o \in \mathbb{R}^{4 \times h_o\times w_o}$.

Since the flat camera image reconstruction is highly ill-posed we utilized a pre-trained diffusion model as our image prior. It is a strong prior for natural images since it is trained on a huge number of samples for the image generation task.

To leverage the diffusion model for our task we need to control its generation process such that we can generate the captured scene from the measurements of the flat camera. To do so we use a ControlNet \cite{zhang2023adding} network $\mathcal{C_\psi}$ which we train for our goal (as presented in \Cref{fig:overview}). 
This network is initialized as a copy of the encoder of the diffusion model UNet with zero convolutions, such that the pre-trained weights performance is not affected and during training non-zero weights are learned for the reconstruction task. 

The input of the control network is the output of the separable transform ($C^o$) and we use the control network loss 
\begin{equation}\label{eq:lc}
l_{\mathcal{C}}=\mathbb{E}_{\mathcal{E}(x), \epsilon \sim N(0,\mathbf{I}), t \sim U(0,T)}\Bigl[ ||\epsilon - \epsilon_\theta(z_t, t, y, \mathcal{C_\psi}(C^o)) ||_2 \Bigr],
\end{equation} 
which is applied for both $\psi$, the set of parameters of the control network $\mathcal{C_\psi}$, and the learned separable transform weights. The diffusion model parameters $\theta$ are pre-trained and fixed.

To guide the training for better results, a separable reconstruction loss term is added to the diffusion conventional loss (\Cref{eq:lc}). This loss is applied to the output of the separable transformation as
\begin{equation}\label{eq:separable_loss}
    l_{sep} = || I - f_{conv}(C^o) ||_2,
\end{equation}
where $f_{conv}$ is a learned $3\times 3$ convolution layer which maps the 4-channels $C^o$ to 3 channels image, and $I$ is the target (GT) image in RGB channels format. We present in \Cref{sec:results:ablations} the improvement and significance of this term.

The diffusion model that we use is a text-guided model for image generation. Thus, we employ this ability to improve the image reconstruction process by giving the model a text description of the captured scene. Giving additional information about the scene content enables the algorithm to have better prior knowledge of the resulting image and reconstruct better images. In this approach the photographer describes the captured scene and this information is input into the reconstruction algorithm. The contribution of the text to the results is presented in \Cref{sec:results:ablations}. 

\subsection{Data Acquisition and Training} \label{sec:method:data}

As we train our DifuzCam model (separable transform and control network) in a supervised way, we need pairs of RGB images with their corresponding measurements of the flat camera.
Simulating the flat camera imaging process to get realistic measurements such that the trained model will generalize to the real world is very hard since the captured measurements are very dependent on camera properties, alignment, calibration, and mask placement. Because of the small feature size (25um), minor movement of the mask will result in totally different measurements. Due to this sensitivity of the system, we obtained using the optical setup real-world measurements that contain all the imaging properties, including space-variant PSF, diffraction, and non-ideal effects which are hard to simulate such as mask manufacturing inaccuracies, dust, image noise, etc. 

To get a large dataset using our flat camera, we captured images from the LAION-aesthetics dataset \cite{schuhmann2022laion}, which were projected on a PC screen of size 34cm $\times$ 34cm at a distance of 60cm from the camera. We used 12ms exposure time and saved the raw Bayer pattern images in 12-bit depth. 
We used LAION-aesthetics \cite{schuhmann2022laion} as it consists of a large amount of high-resolution images and their corresponding textual captions. In total, we captured about 55k images. 500 images were saved for testing while the rest were used for training. To compensate for stray light, a black screen with no image projected on it was captured. In the post-processing stage, the measured black levels were subtracted from the captured measurements.

In addition to the screen images, we also capture real objects to show that we are not restricted only to ``objects on screen''. In this case, we just show the qualitative reconstruction result as we do not have an exact RGB match as in the screen measurement case.

For the pre-trained diffusion model, we used stable-diffusion 2.1 \cite{Rombach_2022_CVPR}. We trained the model for 500k steps on the captured dataset using the looses in \Cref{eq:lc} and \Cref{eq:lldm} and learning rate of $5\cdot 10^{-5}$ with the AdamW optimizer.

\newcommand{\Creal}{figs/compare_real/}
\newcommand{\pfigzoomdog}[1]{\includegraphics[viewport=105 105 230 230, clip, width=0.22\linewidth]{#1}}
\newcommand{\pfigzoomdogb}[1]{\includegraphics[viewport=210 210 460 460, clip, width=0.22\linewidth]{#1}}

\renewcommand{\pfig}[1]{\includegraphics[width=0.22\linewidth]{#1}}
\newcommand{\spaceshift}{2cm}
\begin{figure}
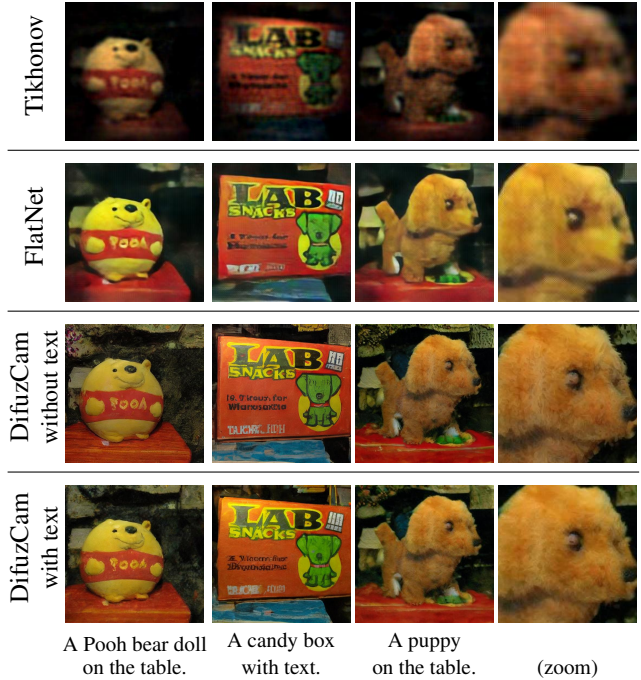

    \setlength\tabcolsep{1pt} 
    \centering
    \begin{tabular}{ccccc}
        \rotatebox{90}{\parbox[c]{\spaceshift}{\centering  Tikhonov}} & \pfig{\Creal rec_9} 
        & \pfig{\Creal rec_7}
        & \pfig{\Creal rec_8}
        & \pfigzoomdog{\Creal rec_8}\\
\hline

        \rotatebox{90}{\parbox[c]{\spaceshift}{\centering  FlatNet}} & \pfig{\Creal fc_9_out} 
        & \pfig{\Creal fc_7_out}
        & \pfig{\Creal fc_8_out}
        & \pfigzoomdog{\Creal fc_8_out}\\
\hline
        \rotatebox{90}{\parbox[c]{\spaceshift}{\centering  DifuzCam \\ \small without text}} 
        & \pfig{\Creal validation_1}
        & \pfig{\Creal validation_2}
        & \pfig{\Creal validation_0}
        & \pfigzoomdogb{\Creal validation_0}\\
\hline
        \rotatebox{90}{\parbox[c]{\spaceshift}{\centering DifuzCam \\ \small with text}} 
        & \pfig{\Creal validation_1t}
        & \pfig{\Creal validation_2t}
        & \pfig{\Creal validation_0t}
        & \pfigzoomdogb{\Creal validation_0t}\\
        
        & \shortstack{\footnotesize A Pooh bear doll \\ \footnotesize on the table.}
        & \shortstack{\footnotesize A candy box \\ \footnotesize with text.} 
        & \shortstack{\footnotesize A puppy \\ \footnotesize on the table.}
        & \shortstack{ \footnotesize (zoom)}\\
        
    \end{tabular}
    \caption{\textbf{Real Scene Results.} Comparison of reconstructions from real scenes measurements with Tikhonov \cite{salman2015flatcam}, FlatNet-T \cite{khan2020flatnet} and Difuzcam (our).}
    \label{fig:flatnet_result_real}
\end{figure}

\renewcommand{\pfig}[1]{\includegraphics[width=0.32\linewidth]{#1}}
\renewcommand{\spaceshift}{2cm}

\renewcommand{\Creal}{figs/real_prototype/}

\section{Experimental Results}
\label{sec:results}

Our DifuzCam method was evaluated using several image quality and text similarity metrics. The results of the proposed reconstruction algorithm were compared to the ground through images using PSNR, SSIM \cite{wang2004image}, and LPIPS \cite{zhang2018perceptual} metrics. Since we also present a text guidance approach, we measured the CLIP score \cite{radford2021learning} compared to a GT text description of the captured scene, a metric that evaluates the similarity of an image to a text. For this metric, we used the latest published and most accurate model (ViT-L/14@336px). On the test dataset that we captured, our method achieved 21.89 in PSNR, 0.541 in SSIM, 0.276 in LPIPS, and a CLIP score of 24.38. \Cref{fig:laion_result} presents examples of our reconstruction results.

\begin{figure}[t]
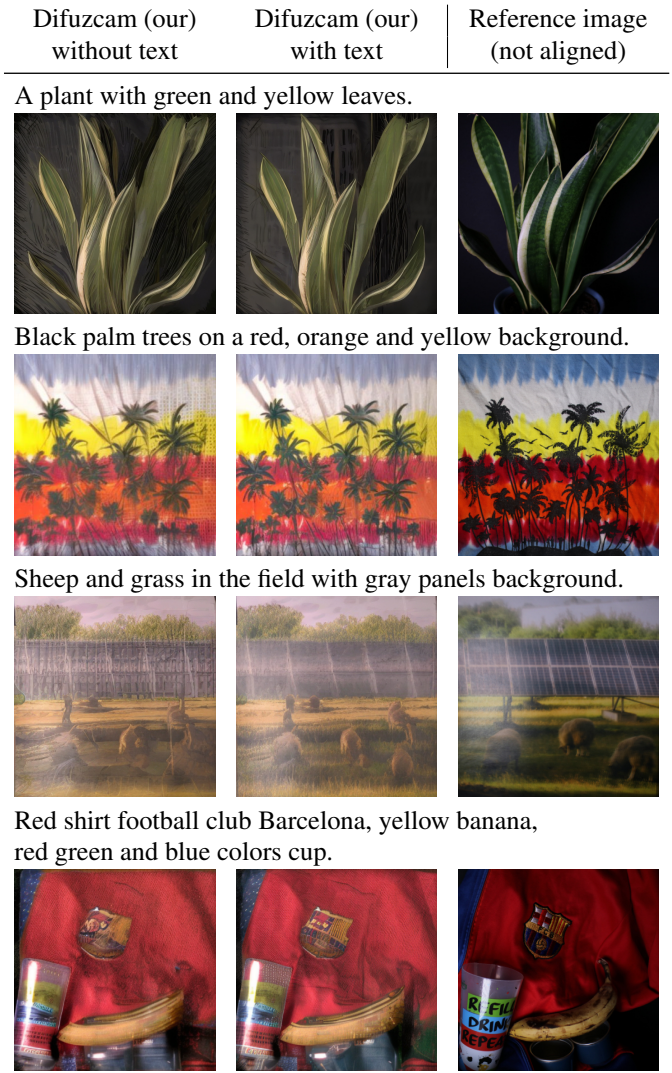

    \setlength\tabcolsep{4pt} 
    \centering
    \begin{tabular}{ccc}
         Difuzcam (our)
        &  Difuzcam (our)
        &  \multicolumn{1}{|c}{Reference image}\\
        without text & with text & \multicolumn{1}{|c}{(not aligned)} \\
        
        \midrule
        \multicolumn{3}{l}{A plant with green and yellow leaves.} \\

        \pfig{\Creal validation_0fc} & \pfig{\Creal validation_0ftc} & \pfig{\Creal IMG_0077} \\

        \multicolumn{3}{l}{Black palm trees on a red, orange and yellow background.} \\

        \pfig{\Creal v82} &
        \pfig{\Creal v82t2} & 
        \pfig{\Creal IMG_0061c2} \\

        \multicolumn{3}{l}{Sheep and grass in the field with gray panels background.} \\
        \pfig{\Creal validation_1c} & 
        \pfig{\Creal validation_1tc} & 
        \pfig{\Creal IMG_0067b} \\
        
        \multicolumn{3}{l}{Red shirt football club Barcelona, yellow banana,} \\
        \multicolumn{3}{l}{red green and blue colors cup.} \\

        \pfig{\Creal validation_5c} & 
        \pfig{\Creal validation_5tc} & 
        \pfig{\Creal IMG_0053c} \\
        
        \bottomrule
    \end{tabular}
    \caption{\textbf{Our Prototype Results.} Real objects were captured with our prototype camera and reconstructed using our proposed method (Difuzcam) with and without text. The reference images were captured with a Canon 80D for visual comparison. Note that they are not accurately aligned with the Difuzcam results.}
    \label{fig:prototype_result_real}
    \vspace{-0.2in}
\end{figure}

To have a fair comparison to previous works, tests should be performed on the same dataset. Since FlatNet \cite{khan2019towardsflatnet} published the dataset of their camera we conducted experiments also on their data to present a valid comparison. Unfortunatelly, we could not compare to \cite{li2023mwdns, pan2022image} as they did not publish their code or model weight for comparison. 
The comparisons to \cite{khan2019towardsflatnet} are presented quantitatively in \Cref{tab:flatnet_results} and visually in \Cref{fig:flatnet_result}. Our method achieved superior results in all evaluated metrics for non-text-guided models. Adding a text to the reconstruction process improves the results in the perceptual and textual similarity metrics. Note that when we just add text without fine-tuning the model to use it a minor degradation in PSNR is observed. When we fine-tune the model to be text-guided we get the best improvement (\Cref{tab:flatnet_results}). 
\Cref{fig:flatnet_result_real} presents results for reconstructions of real scenes (not captured from a screen) that are provided in \cite{khan2020flatnet} and it compares to Flatnet \cite{khan2020flatnet} and Tikhonov \cite{salman2015flatcam}. \Cref{fig:prototype_result_real} presents real scene reconstructions using our prototype camera.

\begin{table}
\renewcommand{\arraystretch}{1.3}
\caption{Results and ablation of our method on dataset captured by the prototype camera. }
\centering
\begin{tabular}{lcccc}
\hline
method & PSNR$\uparrow$ & SSIM$\uparrow$ & LPIPS$\downarrow$ & CLIP$\uparrow$\\
\hline \hline
w/o sep. loss & 9.84&0.175&0.666&23.22
\\
w/o text train & 21.44 & 0.512 & 0.304 & 22.50\\
\textbf{Proposed method} & \textbf{21.58} &\textbf{0.541} & \textbf{0.276} & \textbf{24.38}\\

\hline
\label{tab:laion_ablation_results}

\end{tabular}
\vspace{-0.15in}
\end{table}

\subsection{Implementation details}\label{sec:results:impldeatail}
In the Laion dataset, the given text captions for the images are not always accurate or relevant. We acknowledge that this noise in the data might harm the results we get in the training process when using these text captions.
Since we identified that these inaccuracies might be critical in the tests, we manually checked the test dataset captions to verify the accuracy and correctness of the data. This verification is very important for the text guidance reconstruction results and also for the textual CLIP score evaluation we made.
Despite the potential disadvantage of training on incorrect captions, we did not manually verify the training dataset since it is not feasible to manually check such a very large dataset.

To compare our results to FlatNet \cite{khan2019towardsflatnet}, we trained our method on their published dataset which consists of 10k images for training and 100 for testing. This data does not contain captions to the images. Thus, to train our method with text guidance on this data we use a large language model (LLM) for the auto image captioning process. We used llava1.5 \cite{liu2024visual} LLM and generated captions for all the images in the data. Here we also might have the problem of incorrect captions, which is also known as LLM hallucinations. Also in this case, the test samples captions were manually verified due to the high importance of the test captions' correctness. For this data we trained the model for 700k steps with a similar optimizer setup to what we mentioned in \Cref{sec:method:data}.
We used the Allied Vision 1800 U-500 board-level camera with a pixel size of  2.2um and 5 megapixels overall for the prototype camera.

\renewcommand{\pfig}[1]{\includegraphics[width=0.4\linewidth]{#1}}
\newcommand{\pfigzoomA}[1]{\includegraphics[viewport=180 100 380 300, clip, width=0.36\linewidth]{#1}}
\newcommand{\pfigzoomB}[1]{\includegraphics[viewport=50 100 450 500, clip, width=0.36\linewidth]{#1}}

\newcommand{\Emix}{figs/laion/mix/}
\begin{figure}[t]
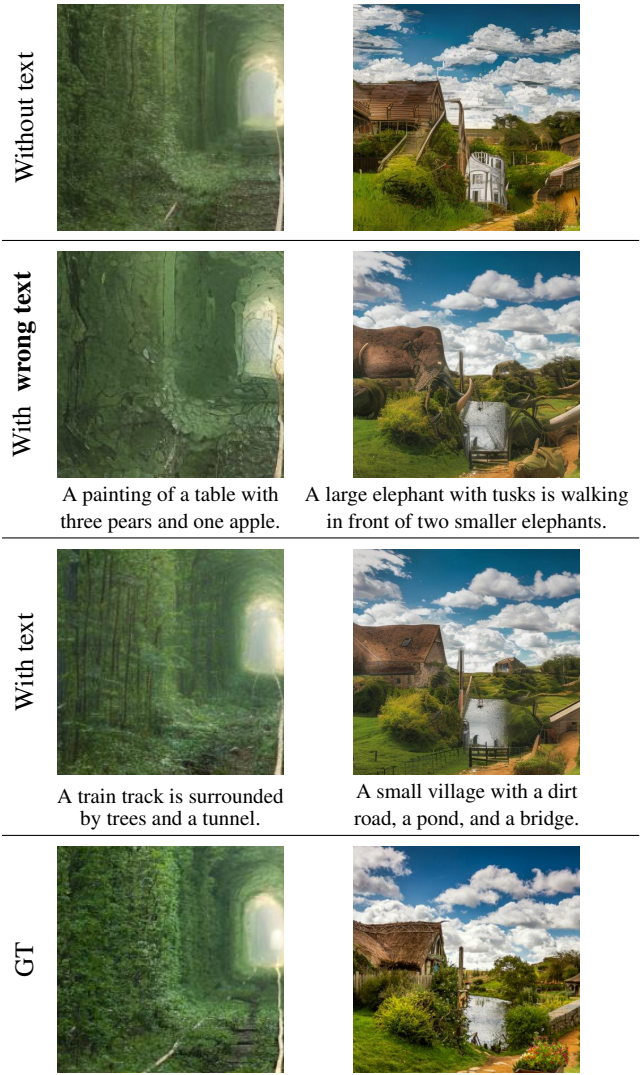


    \setlength\tabcolsep{4pt} 
    \centering
    \begin{tabular}{ccc}
        \rotatebox{90}{\parbox[c]{3cm}{\centering  Without text}} & \pfigzoomA{\Eournc validation_14} & \pfigzoomB{\Eournc validation_34}
        \\
\hline \vspace{-8pt} \\
        \rotatebox{90}{\parbox[c]{3cm}{\centering  With \textbf{ wrong text}}} & \pfigzoomA{\Emix validation_14_mix} & \pfigzoomB{\Emix validation_34_mix}
        \\

        & \shortstack{\footnotesize A painting of a table with \\ \footnotesize three pears and one apple.}
        & {\footnotesize \shortstack{ A large elephant with tusks is walking \\ in front of two smaller elephants.}}
        \\
        \hline \vspace{-8pt} \\
        \rotatebox{90}{\parbox[c]{3cm}{\centering With text}} & \pfigzoomA{\Eour validation_14} & \pfigzoomB{\Eour validation_34}
        \\

        & \shortstack{\footnotesize A train track is surrounded \\ \footnotesize by trees and a tunnel.}
        & \shortstack{\footnotesize A small village with a dirt \\ \footnotesize road, a pond, and a bridge.}
        \\

        \hline \vspace{-8pt} \\
        \rotatebox{90}{\parbox[c]{3cm}{\centering GT}} & \pfigzoomA{\Egt targets_14} & \pfigzoomB{\Egt targets_34}\\

    \end{tabular}
    \caption{\textbf{Ablation Results.} Showing the contribution of the text to the reconstruction. Without text input, the reconstruction is driven by the flat camera measurements only. With text guidance, the reconstructed image details are more similar to the true captured scene. Yet, when a wrong image caption is given, the reconstructed details and high frequencies might be wrong and less compatible with the scene. } 
    
    \label{fig:laion_ablation}
    \vspace{-0.15in}
\end{figure}

\subsection{Ablations}\label{sec:results:ablations}
We present ablation results in \Cref{tab:laion_ablation_results} and \Cref{fig:laion_ablation}. First of all, it is noticeable from \Cref{tab:laion_ablation_results} that without our proposed separable loss the reconstructed images are not similar to the target image. We observe that the reconstructed images contain the information of the text caption, according to the high CLIP score, but do not succeed in extracting additional information from the camera measurements for the reconstruction process, i.e., the reconstructions become independent of the input camera measurements. When we do use the separable loss, the measurements are taken into account. Adding text information as input improves the reconstruction even further, compared to the non-text-guided model. The visual ablation images in \Cref{fig:laion_ablation} show that the text captions contribute to the high frequency details in the reconstructed images. When we supply a text caption, the reconstructed image details are aligned with the text. This is noticeable also when a wrong text caption is provided. For example, the reconstruction gains a painting style when the caption mentions a painting and elephant shapes are visible when elephants are described in the caption.

\section{Conclusion}
\label{sec:conclusion}

A novel method for image reconstruction from flat camera measurements was presented, achieving high quality reconstructions, with and without text guidance. The method leverages the strong capabilities of a pre-trained diffusion model for image prior. Such an approach can be integrated into other imaging systems to improve the reconstructions. 
Even though we get perceptually pleasant reconstructed images, one may notice minor inaccuracies in the reconstructed fine details compared to the ground-through image. Since the imaging method at hand is highly ill-posed, the model learns to generate the missing details, which have been lost in the acquisition process. We therefore do not consider this as a problem but rather a property of the method that improves the reconstruction quality. As we have demonstrated in this work, our approach presents state-of-the-art results compared to previous results and significantly improves the reconstruction abilities of flat cameras.

\bibliographystyle{ieee_fullname} 
\bibliography{references}


\end{document}